\documentclass[runningheads]{llncs}
\usepackage[T1]{fontenc}
\usepackage{float}
\usepackage[frozencache,cachedir=minted]{minted}
\setminted{style=trac}
\usepackage{microtype}
\usepackage{float}
\usepackage{graphicx}
\usepackage{xcolor}
\usepackage[hidelinks]{hyperref}
\usepackage{amsmath}
\usepackage{amssymb}
\usepackage[capitalise]{cleveref}
\usepackage{multirow}
\usepackage{makecell}
\usepackage{tabularx}
\usepackage{booktabs}
\usepackage{fancyvrb}
\usepackage{todonotes}
\usepackage{fancyhdr}

\fancypagestyle{firstpage}{
	\fancyhf{}
	
	\fancyfoot[L]{\footnotesize This version of the contribution has been accepted for publication, after peer review but is not the Version of Record and does not reflect post-acceptance improvements, or any corrections. The Version of Record is available online at: \url{http://dx.doi.org/10.1007/978-3-032-09527-5_15}. Use of this Accepted Version is subject to the publisher’s Accepted Manuscript terms of use \url{https://www.springernature.com/gp/open-research/policies/accepted-manuscript-terms}}
}

\definecolor{fncolor}{RGB}{0, 1, 73}
\def\FN#1{{\color{fncolor} \texttt{#1}}}
\definecolor{fnsetcolor}{RGB}{137, 83, 148}
\def\FNSET#1{{\color{fnsetcolor} \boldmath{#1}}}
\usepackage{color}

\begin{document}
\title{GRASP: Generic Reasoning And SPARQL Generation across Knowledge Graphs}

\titlerunning{GRASP}
%
\author{Sebastian Walter\inst{1}\orcidID{0009-0006-2613-3209} \and Hannah Bast\inst{1}\orcidID{0000-0003-1213-6776}}
\authorrunning{S. Walter and H. Bast}
%
\institute{University of Freiburg, 79100 Freiburg im Breisgau, Germany
\email{{swalter,bast}@cs.uni-freiburg.de}}

\maketitle              

\thispagestyle{firstpage}

\vspace{-4mm}
\begin{abstract}
We propose a new approach for generating SPARQL queries on RDF knowledge graphs from natural language questions or keyword queries, using a large language model. Our approach does not require fine-tuning. Instead, it uses the language model to explore the knowledge graph by strategically executing SPARQL queries and searching for relevant IRIs and literals. We evaluate our approach on a variety of benchmarks (for knowledge graphs of different kinds and sizes) and language models (of different scales and types, commercial as well as open-source) and compare it with existing approaches. On Wikidata we reach state-of-the-art results on multiple benchmarks, despite the zero-shot setting. On Freebase we come close to the best few-shot methods. On other, less commonly evaluated knowledge graphs and benchmarks our approach also performs well overall. We conduct several additional studies, like comparing different ways of searching the graphs, incorporating a feedback mechanism, or making use of few-shot examples.
\end{abstract}

\vspace{-6mm}
\def\B{$\,$B\ }
\section{Introduction}


More and more datasets are available as RDF, one of the standard data models for knowledge graphs.\footnote{Whenever we speak of knowledge graphs, we mean RDF knowledge graphs.} Early examples are DBpedia \cite{dbpedia} and Freebase \cite{freebase}, followed by Wikidata \cite{wikidata}, and more recently UniProt \cite{uniprot-rdf}, OpenStreetMap \cite{osm2rdf} (OSM), and DBLP \cite{dblptgdk}.
The standard query language for RDF data is SPARQL, which can be powerful but also very challenging for multiple reasons. For example, consider the following question on Wikidata:
``Who was the Governor of Ohio by the end of 2011?''.\footnote{This is a question very similar to one from our benchmarks in \cref{subsec:bemcharksandmetric}.}
Here is a SPARQL query that answers the question:\footnote{See \href{https://www.wikidata.org/wiki/EntitySchema:E49}{www.wikidata.org} for the PREFIX definitions for \emph{p:}, \emph{ps:}, \emph{wd:}, \emph{pq:}, and \emph{xsd:}.}

\vspace{-1mm}
\begin{minted}[escapeinside=``]{sparql}
SELECT ?result WHERE {
  ?result p:P39 ?m .
  ?m ps:P39 wd:Q17989863 ; pq:P580 ?start .
  OPTIONAL { ?m pq:P582 ?end . }
  FILTER(?start <= "2011-12-31"^^xsd:date &&
         (!BOUND(?end) `||` ?end >= "2011-12-31"^^xsd:date))
}
\end{minted}
Let us go through the many challenges of formulating this SPARQL query:\\[1mm]
1. One has to know the basic syntax of SPARQL.\\[0.5mm]
2. One has to know that in Wikidata, the information about which person holds which position is modeled by the property \emph{P39} (``position held``).\\[0.5mm]
3. One has to know how the quaternary information about the position (person, position, start, end) is modeled in Wikidata and that the IRIs of the respective predicates are \emph{p:P39}, \emph{ps:P39}, \emph{pq:580}, and \emph{pq:P582}.\\[0.5mm]
4. One has to know that there is an own entity for ``governor of Ohio'' (as opposed to separate entities for ``governor'' and ``Ohio'') and that its IRI is \emph{wd:Q17989863}.\\[0.5mm]
5. One has to realize that the person might still be in office, so that the end date has to be OPTIONAL.\\[0.5mm]
6. One has to translate ``in office at the end of 2011'' to a logical expression that states that the term started on or before the last day of 2011, and ended after that or is still ongoing.\\[1mm]
%
Based on our own extensive experience with SPARQL users, let us briefly discuss how a casual SPARQL user, an expert SPARQL user, and an unassisted large language model (LLM) will perform at this task.
All three of them know 1.
Both the expert and the LLM will know 2 and 3 because that is a common construct in Wikidata queries.
Both the casual and the expert user will have to do some research to find out how ``governor of Ohio'' is expressed in Wikidata and with which IRI.\footnote{For this, an autocompletion mechanism is typically used; see \cite{qleverautocompletion} for an overview.}
An unassisted LLM will hallucinate an IRI and almost certainly get it wrong.
The casual user might miss 5, the expert user will know this, for the LLM it can go either way.
The casual and the expert user will eventually find out 6, but with some trial and error.
The unassisted LLM will often get it wrong without feedback.
Also note that in practice, LLMs act probabilistically, and the generated query varies widely between attempts, even for the exact same prompt.
For the question above, the query produced by an unassisted LLM is never 100\% correct, but sometimes more and sometimes less wrong.

More generally, an LLM requires two kinds of assistance to be able to formulate SPARQL queries for arbitrary given questions: (1) it needs help to find the right IRIs and literals, and (2) it needs help with the finer points of formulating the query. Let us briefly discuss the challenges of both of these.\\[1mm]
\begin{figure}[h!]
\centering
\includegraphics[width=\textwidth]{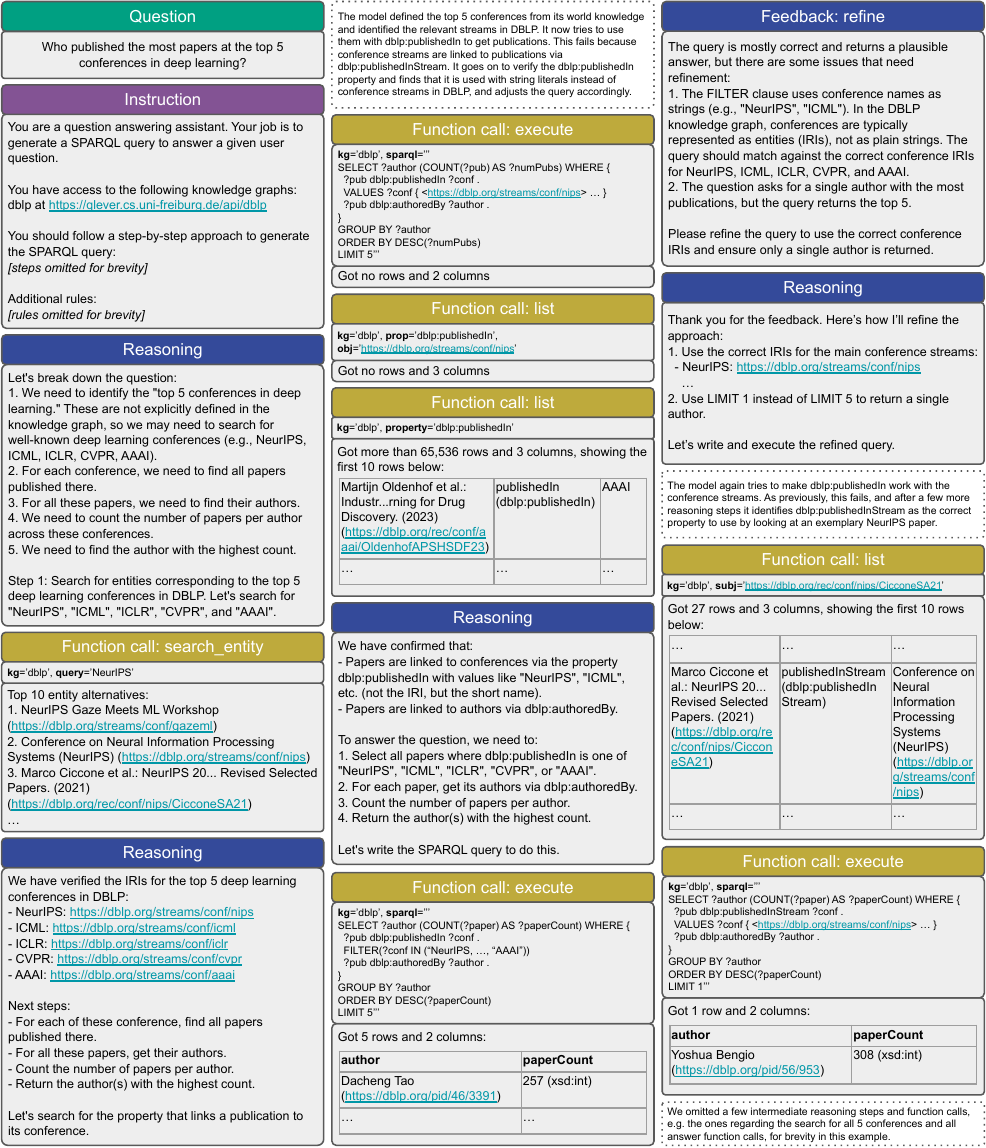}
\caption{Overview over GRASP with feedback using the question ``Who published the most papers at the top 5 conferences in deep learning?'' as a running example.}
\label{fig:grasp}
\vspace{-4mm}
\end{figure}

Issue (1) is hard because of the sheer size of many knowledge graphs. For example, Wikidata contains 1.2 \B distinct literals and 2.2 \B subjects. It would require a gargantuan
LLM to memorize these and even if that was achieved, it would be a waste of resources. A more reasonable approach would be to either post-process the query by replacing hallucinated IRIs and literals by the matching ones or to store the IRIs and literals externally and provide a means for the LLM to query them. Regarding (2), one could fine-tune the LLM to learn frequently used query patterns and then hope for the best. Or one could provide some feedback mechanism to let the LLM reason and let it know whether it is on the right track.

We explore and evaluate a variety of approaches in this paper. Our goal is an approach that works for \emph{any} given knowledge graph without the need for fine-tuning, just by dynamically exploring the graph. We do this by providing the LLM with a fixed set of {\color{fncolor}\textbf{\emph{functions}}}. These are independent of the graph; all that differs is the SPARQL endpoint that the LLM queries. 
We call our approach \emph{GRASP}, which stands for \emph{Generic Reasoning And SPARQL Generation across Knowledge Graphs}.
To prove its versatility, we evaluate our approach on a variety of knowledge graphs, including those named at the beginning of the introduction. To be the best of our knowledge, GRASP is the first approach of this kind and the first to be evaluated on such a wide variety of knowledge graphs.

\vspace{-2mm}
\subsubsection{Problem Formulation} We consider the following \emph{zero-shot SPARQL QA problem}: Given an arbitrary natural-language question and a SPARQL endpoint for an arbitrary RDF knowledge graph, compute the SPARQL query that corresponds to the natural-language question, or conclude that there is no such query on that graph. There is no training data, and fine-tuning of the model on the graph is not allowed.
However, it is permissible to pre-compute index data structures for searching the set of IRIs and literals in the graph efficiently.
\footnote{Specifically, for each of the knowledge graphs we investigate in this paper, we have written a SPARQL query that extracts the label and optional additional information for each IRI; see \cref{subsec:indices}. This is query-independent and the SPARQL query could, in principle, also be found by our approach itself.}

\def\REPRO{\href{https://github.com/ad-freiburg/grasp}{github.com/ad-freiburg/grasp}}
\vspace{-2mm}
\subsubsection{Contributions} Our core contributions are as follows:\\[1mm]
1. We present GRASP, the first approach to the zero-shot SPARQL QA problem that, in principle, works for arbitrary RDF knowledge graphs.\\[0.5mm]
2. We evaluate GRASP on a variety of knowledge graphs (DBLP, DBpedia, Freebase, ORKG, OSM, UniProt, Wikidata).
On Freebase and Wikidata, we compare GRASP extensively with existing approaches.
\\[0.5mm]
%
3. We conduct several studies to examine how different components of GRASP, such as model choice, the set of functions, the use of few-shot examples or feedback, affect its performance.\\[0.5mm]
4. We provide full reproducibility materials via {\REPRO}, in particular: our code and knowledge graph indices, all benchmarks and model outputs, and an evaluation web app for interactively exploring model results.




\section{Related Work}


Our zero-shot SPARQL QA problem is closely related to the more general problem of knowledge graph question answering (KGQA), which allows answers to questions to be found through other ways than by generating SPARQL queries. We divide KGQA methods into three categories.

In the first category are methods that fine-tune on a particular benchmark for a particular knowledge graph. Examples for such methods are DeCAF \cite{decaf}, RoG \cite{rog}, ChatKBQA \cite{chatkbqa}, and many more \cite{pangu,flexkbqa,rgrkbqa,rwt,wikisp,symagent,lightprof}. State-of-the-art on KGQA benchmarks is typically achieved by such methods because they can adapt to the peculiarities of the benchmark and knowledge graph and exploit similarities between training and test sets.

In the second category are methods that do not fine-tune on a particular benchmark dataset but still have access to benchmark specific information in-context during inference, most often in the form of exemplary question-SPARQL pairs or input-output examples for submodules. Examples in this category are KB-BINDER \cite{kbbinder}, Pangu \cite{pangu}, DoG \cite{dog}, KELDaR \cite{keldar} and Interactive-KBQA \cite{interactivekbqa}. They achieve good performance, sometimes even close to fine-tuned methods, but they either use a large number of examples (100 for KB-BINDER, 1000 for Pangu, 10/10/1 for different modules in DoG, 40/6/10/4 for different modules in KELDaR) or few extensive ones (full generation processes for Interactive-KBQA). Interactive-KBQA is similar to our approach, but with different functions, some optimized for specific knowledge graphs (e.g. \emph{SearchGraphPatterns} for CVT structures in Freebase), and tailored instructions with hints for each graph.

In the third category are zero-shot methods that work without fine-tuning or examples. StructGPT \cite{structgpt} finds answer entities starting from a given topic entity by moving along edges and stopping once it assesses the newly discovered entities suitable to answer the question. Similarly, ToG \cite{tog} explores the knowledge graph starting from topic entities along edges and nodes, and always keeps track of the overall best paths so far.
SPINACH \cite{spinach} is similar to our approach and equips a LLM with functions to search in and execute queries over a knowledge graph. However, they only focus on Wikidata and use the Wikidata API to implement their search functionality, limiting transferability to other knowledge graphs. They also structure the generation process differently, with separate prompts for different subtasks, like language detection, action generation, output generation, SPARQL extraction, and the pruning of Wikidata entries, for which they use two Wikidata-specific examples in context. In contrast, our approach works with a single instruction that is the same across all knowledge graphs and a single-context continuous generation process from question to SPARQL query.


\section{Approach}

We introduce our overall approach as well as its different components in the following. For all SPARQL-related functionality, such as retrieving data from and executing queries over knowledge graphs, we use the public SPARQL endpoints at \href{https://qlever.cs.uni-freiburg.de}{qlever.cs.uni-freiburg.de} \cite{qlever} because of QLever's superior performance for hard queries compared to alternatives like Blazegraph or Virtuoso.

\subsection{GRASP}

With GRASP, the generation of SPARQL queries relies on a LLM that interacts with a knowledge graph autonomously by interleaving function calls to search or query the knowledge graph with reasoning about previous and future steps. Note, that this is in style similar to ReAct prompting \cite{react}, though we also allow multiple successive reasoning steps or function calls.
Starting point of a generation is the GRASP instruction, which contains the name and endpoint of all available knowledge graphs, details about the step-by-step approach, and additional rules to be followed. Note, that this instruction is generic and does not contain any graph-specific information other than their names and endpoints. 
Optionally, few-shot examples (see \cref{subsec:fsl}) and a feedback mechanism for refining the output (see \cref{subsec:feedback}) can be added. See \cref{fig:grasp} for an overview over GRASP's approach by means of an example for DBLP.

\subsection{Functions}\label{sec:functions}

In the following we define all functions that we use for interaction between model and knowledge graph. Each function has a \FN{kg} parameter for dynamically specifying the knowledge graph.
The model can thus choose the most suitable graph itself. This mechanism even allows federated queries across multiple graphs. We give every function a three-letter mnemonic and denote optional and enum arguments with \FN{?} and \FN{|} respectively.\\[1mm]
\FN{ANS: answer(kg: str, sparql: str, answer: str)}\\
Provides the final SPARQL query, the knowledge graph to execute it on, and a human-readable answer to the question. Calling this function stops the generation process.\\[1mm]
\FN{CAN: cancel(expl: str, best\_attempt: \{sparql: str, kg: str\}?)}\\
If no satisfactory SPARQL query can be found, calling this function allows to give an explanation for that and an optional best attempt at the SPARQL query. It also stops the generation process.\\[1mm]
\FN{EXE: execute(kg: str, sparql: str)}\\
Execute a SPARQL query over the knowledge graph and retrieve its results as a table in the case of success, and an error message otherwise. If the result table has more than 10 rows or columns, we only show the first five and last five rows or columns respectively. The client-side query timeout is set to 60 seconds.\footnote{We consider 10 rows and 10 columns to be enough to get a sense of the query result. Larger timeouts didn't improve results significantly.}\\[1mm]
%
\FN{LST: list(kg: str, subj: str?, prop: str?, obj: str?)}\\
List triples from the knowledge graph according to the given constraints on the subject, property, and object positions. It returns at most 10 triples and makes sure that the returned ones are relevant and diverse.\footnote{A larger number of triples costs more tokens, but did not improve results significantly.}
For example, we prefer entities and properties with high scores and avoid multiple occurrences of the same entities and properties.\\[1mm]
%
\FN{SEN: search\_entity(kg: str, query: str)}\\
Search for entities in the specified knowledge graph with the query. Uses a prefix-keyword-index internally.\\[1mm]
\FN{SPR: search\_property(kg: str, query: str)}\\
Search for properties in the specified knowledge graph with the query. Uses a similarity index internally.\\[1mm]
\FN{SPE: search\_property\_of\_entity(kg: str, query: str, ent: str)}\\
Same as \FN{SPR}, but restricts the search to properties that exist for the given entity in the knowledge graph.\\[1mm]
\FN{SOP: search\_object\_of\_property(kg: str, query: str, prop: str)}\\
Similar to \FN{SEN}, but restricts the search to entities and literals that exist for the given property in the knowledge graph at the object position.\\[1mm]
\FN{SAC: search\_autocomplete(kg: str, query: str, sparql: str)}\\
Search for items in the knowledge graph with the query in a context-sensitive way. The given SPARQL query must be a SELECT query with a variable \emph{?search} in its body. The search is then restricted to items that fit at the variable's position.\\[1mm]
\FN{SCN: search\_constrained(kg: str, query: str, pos: subj|prop|obj,\\constraints: \{subj: str?, prop: str?, obj: str?\})}\\
Search for triple items at the specified position in the knowledge graph with the query. If there are further constraints specified, they are used to limit the search space to matching triples accordingly.\\[1mm]
All search functions return the top 10 search results, which we consider to be enough in most scenarios.\\[1mm]
\FN{FSE: find\_similar\_examples(kg: str, question: str)}\\
Find question-SPARQL-pairs over the specified knowledge graph that have questions similar to the given one. It returns the three most similar pairs based on a pre-computed similarity index. In this work, the index is built from the training set of the respective benchmark we evaluate on, if available.\\[0.5mm]
%
\FN{FEX: find\_examples(kg: str)}\\
Like \FN{FSE} but the returned examples are selected randomly.\\[1mm]
Note, that the two functions above are only used for few-shot evaluations.


\vspace{-3mm}
\subsection{Function Sets}\label{sec:function_sets}

We evaluate different variants of our approach, where each variant has a particular subset of the functions from \cref{sec:functions} available. We call such a subset a \FNSET{\textbf{\emph{function set}}}.
We start with the basic function set \FNSET{$B = \{\FN{ANS}, \FN{CAN}, \FN{EXE}\}$}, which allows the model to interact with the knowledge graph solely by executing SPARQL queries.
For the other functions sets, we build upon \FNSET{$B$} and add \FN{LST} to enable the model to retrieve RDF triples, and one or more search functions to enable the model to search the knowledge graph. We use \FN{SEN} and \FN{SPR} for the function set \FNSET{$S$}, \FN{SAC} for the function set \FNSET{$S_A$} and \FN{SCN} for the function set \FNSET{$S_C$}. In theory, the model can perform the most sophisticated searches with \FN{SAC}, followed by \FN{SCN}, and then the search functions in \FNSET{$S$}. The first two can dynamically constrain searches to subgraphs, with \FN{SAC} providing even more flexibility than \FN{SCN} because it allows for full-blown SPARQL queries as constraining element rather than a single RDF triple. However, during initial experiments with these functions, we noticed that the model often struggles to use them properly, or only uses them in their most basic form for unconstrained search. Therefore, we introduce the \FNSET{$S_E$} function set to strike a balance between ease of use and functionality. This function set builds upon the function set \FNSET{$S$} but contains additional specialized functions for the two most common types of searches a human expert performs when writing SPARQL queries: \FN{SPE} to look for properties of a given entity, and \FN{SOP} to look for entities and literals reachable by a given property. For few-shot evaluations we also add one of \FN{FSE} or \FN{FEX} to each function set. See \cref{subsec:fsl} for more details. See \cref{tab:fnsets} for an overview over all function sets.

\begin{table}
\centering
\caption{The function sets and which functions they include. The \FN{FSE} and \FN{FEX}  functions are only added for few-shot evaluations.}
\begin{tabular}{lcccccccccccc}
\toprule
\textbf{Set} & \FN{\textbf{ANS}} & \FN{\textbf{CAN}} & \FN{\textbf{EXE}} & \FN{\textbf{LST}} & \FN{\textbf{SEN}} & \FN{\textbf{SPR}} & \FN{\textbf{SPE}} & \FN{\textbf{SOP}} & \FN{\textbf{SCN}} & \FN{\textbf{SAC}} & \FN{\textbf{FSE}} & \FN{\textbf{FEX}} \\ \midrule
\FNSET{\textbf{$B$}} & \checkmark & \checkmark & \checkmark & & & & & & & & (\checkmark) & (\checkmark) \\
\FNSET{\textbf{$S$}} & \checkmark & \checkmark & \checkmark & \checkmark & \checkmark & \checkmark & & & & & (\checkmark) & (\checkmark)
\\
\FNSET{\textbf{$S_E$}} & \checkmark & \checkmark & \checkmark & \checkmark & \checkmark & \checkmark & \checkmark & \checkmark & & & (\checkmark) & (\checkmark)
\\
\FNSET{\textbf{$S_C$}} & \checkmark & \checkmark & \checkmark & \checkmark & & & & & \checkmark & & (\checkmark) & (\checkmark)
\\
\FNSET{\textbf{$S_A$}} & \checkmark & \checkmark & \checkmark & \checkmark & & & & & & \checkmark & (\checkmark) & (\checkmark)
\end{tabular}

\label{tab:fnsets}
\end{table}

\vspace{-3mm}
\subsection{Search Indices}
\label{subsec:indices}

The search functions are all based on pre-computed index data structures, which we describe in the following. See \cref{tab:kgindices} for details about all used indices.

\subsubsection{Data} The data for the index data structures comes from two SPARQL queries per knowledge graph, one for entities and one for properties. They compute one row per item with five columns each: IRI, label, score, synonyms, and additional information. The IRI is unique and besides it only the label is required. The rows are ordered by descending score, with a higher score corresponding to more popular or more commonly used items.
\footnote{In the indices the score is used to resolve tie-breaks between equally ranked items.} 
Synonyms define alternative labels that are indexed like the main label.\footnote{When querying an index, labels and synonyms of each IRI are ranked separately, but the highest-ranked one determines the IRI’s final rank.} The additional information is not indexed but can help to disambiguate similarly labeled items. See \cref{tab:wikidataentities} for example data.\footnote{See the corresponding SPARQL query at \href{https://qlever.cs.uni-freiburg.de/wikidata/InnqDm}{qlever.cs.uni-freiburg.de/wikidata/InnqDm}.}


\begin{table}
\centering
\caption{Index data for four example Wikidata entities.}
\begin{tabularx}{\textwidth}{l p{1.5cm} c X p{3.5cm}}
\toprule
\textbf{IRI} & \textbf{Label} & \textbf{Score} & \textbf{Synonyms} & \textbf{Infos} \\
\midrule
\href{http://www.wikidata.org/entity/Q9047}{wd:Q9047} & Gottfried Wilhelm Leibniz & 202 & Gottfried Wilhelm; Leibniz; Gottfried Wilhelm von Leibniz; Leibnitz; Freiherr Gottfried Wilhelm von Leibniz; ... & archivist; jurist; zoologist; philosopher of law; writer; ... \\
\href{http://www.wikidata.org/entity/Q5202}{wd:Q5202} & Almazán & 167 & Almazan & municipality in the province of Soria, Castile and León, Spain \\
\href{http://www.wikidata.org/entity/Q12117}{wd:Q12117} & cereal grain & 147 & cereal; cereals & fruits (grains) of cereal crops used for food and agricultural products \\
\href{http://www.wikidata.org/entity/Q9950}{wd:Q9950} & Q & 142 & q & letter of Latin alphabet \\
\end{tabularx}
\label{tab:wikidataentities}
\end{table}

\vspace{-8mm}
\subsubsection{Prefix-keyword index} Our prefix-keyword index scores and ranks entities by the number of exact and prefix matches between the entity keywords and a keyword query. For example, given the query ``Albert E'' and entities \{\emph{Peter Falk}, \emph{Carlos Alberto}, \emph{Albert Einstein}, \emph{Albert Finney}\} our index will return the ranked list [\emph{Albert Einstein}, \emph{Albert Finney}, \emph{Carlos Alberto}].\footnote{\emph{Albert Einstein} has one exact and one prefix match, \emph{Albert Finney} one exact match, and \emph{Carlos Alberto} one prefix match. \emph{Peter Falk} is excluded because it has no match.} 

\subsubsection{Similarity index} Our similarity index scores and ranks properties by the cosine-similarity between search query embedding and property embeddings. For vector search on the pre-computed property embeddings we use FAISS \cite{faiss}. For our embedding model we use \href{https://huggingface.co/mixedbread-ai/mxbai-embed-large-v1}{\emph{mxbai-embed-large-v1}} \cite{mxbai} within the Sentence Transformers library \cite{sentencetransformers}.

\begin{table}[h]
\centering
\caption{Details about the size of our indices for various knowledge graphs. We use prefix-keyword indices for entities and similarity indices for properties. \\\# refers to the number of indexed items, while Data and Index refer to the disk space occupied by the data from the SPARQL queries and index data structures respectively. * The Open Research Knowledge Graph (ORKG) indices are based on an old version from 2023 \cite{orkg2023}, that is needed to evaluate on SciQA \cite{sciqa}.}
\begin{tabular}{lrrrrrr}
\toprule
\multirow{2}{*}{\textbf{KG}} &
\multicolumn{3}{c}{\textbf{Entities}} &
\multicolumn{3}{c}{\textbf{Properties}} \\
 & \multicolumn{1}{c}{\#} & \multicolumn{1}{c}{Data} & \multicolumn{1}{c}{Index}
& \multicolumn{1}{c}{\#} & \multicolumn{1}{c}{Data} & \multicolumn{1}{c}{Index} \\
\midrule
DBLP \cite{dblptgdk}         & 141M  & 13GB   & 11GB  & 67    & 12kB   & 280kB \\
DBpedia   \cite{dbpedia} & 19M   & 3GB    & 906MB & 54k   & 3MB    & 213MB \\
Freebase \cite{freebase} & 47M   & 6GB    & 2GB   & 21k   & 3MB    & 82MB  \\
*ORKG    \cite{orkg}      & 180k  & 16MB   & 8MB   & 7k    & 440kB  & 27MB  \\
OSM \cite{osm2rdf}    & 106M  & 7GB    & 4GB   & 101K  & 8MB    & 397MB \\
UniProt \cite{uniprot-rdf}   & 4M    & 343MB  & 317MB & 237   & 24kB   & 960kB \\
Wikidata \cite{wikidata} & 85M   & 12GB   & 6GB   & 12k   & 2MB    & 134MB \\
\end{tabular}
\label{tab:kgindices}
\end{table}

\vspace{-4mm}
\subsection{Few-shot Examples}
\label{subsec:fsl}

Our focus with GRASP is zero-shot inference, but we are still interested in whether providing a few question-SPARQL examples in context can improve performance. For a given knowledge graph and question, the few-shot examples are retrieved from the training set of the corresponding benchmark (if one exists) at the beginning of the generation process, either randomly or by similarity to the question. This is implemented via the functions \FN{FSE} and \FN{FEX}.

\vspace{-2mm}
\subsection{Feedback}
\label{subsec:feedback}

We also investigate how providing feedback on calls of \FN{ANS} or \FN{CAN} can help the model produce better outputs. We do this especially to improve rule following for the rules specified in the GRASP instruction, because this typically suffers in long generation traces. Importantly, the feedback model is the same as the SPARQL generating one but receives only the \FN{ANS} or \FN{CAN} arguments and the SPARQL rules from the GRASP instruction as input, not the full generation history. The feedback consists of a status \FN{done|refine|retry} and  a general feedback message. If the status is \FN{done}, we stop ultimately. Otherwise, we add the feedback to the history and continue the generation process until the next \FN{ANS} or \FN{CAN} call. We allow at most two intermediate feedback loops.

\vspace{-2mm}
\section{Evaluation}

In this section, we present our evaluation results. We start by describing the evaluation setup, such as the benchmarks and metric used, as well as the different model configurations we look at. Unless stated otherwise, we use the \FNSET{$S_E$} function set. We call the variant of our approach with feedback enabled GRASP-F.

\subsection{Benchmarks and Metric}
\label{subsec:bemcharksandmetric}

In the following, we introduce the main benchmark datasets we evaluate on in this work. We chose them because they are either widely used in the research community or very relevant to our work. We also evaluate our approach on more knowledge graphs and benchmarks, see \cref{subsec:various_studies} and \cref{tab:component_kgs} for details.

\vspace{-2mm}
\subsubsection{Benchmarks} WebQuestionsSP (WQSP) annotates the WebQuestions dataset \cite{webquestions} with SPARQL queries over Freebase \cite{webqsp}. ComplexWebQuestions (CWQ) extends WQSP with automatically generated, more complex SPARQL queries, and annotates them with corresponding natural language questions \cite{cwq}. QALD-7 \cite{qald7} is the first QALD edition involving Wikidata. QALD-10 \cite{qald10} then focuses on Wikidata, with a new large test set with real-world questions and expert-annotated SPARQL queries. SPINACH is created from real-world SPARQL queries in Wikidata's ``Request a query'' forum, annotated with natural language questions \cite{spinach}. WikiWebQuestions (WWQ) adapts WQSP to Wikidata \cite{wikisp}.

\vspace{-2mm}
\subsubsection{Metric}
All models are evaluated using the F\textsubscript{1}-score averaged across samples. We adopt SPINACH’s modification, which permits additional columns (e.g., labels) in the predicted results without penalty.%
\footnote{For results exceeding 1,024 rows, we revert to the standard exact F\textsubscript{1}-score due to the computational cost of row-wise assignment.} Samples with empty groundtruth are excluded. If the predicted query is an ASK query and the groundtruth is a SELECT query (or vice versa), we assign a score of 1 if their results are semantically equivalent.
%
To save compute and money, we evaluate on 200 random samples for each benchmark, which for QALD-7 and SPINACH is still the whole test set. For the component studies in \cref{subsec:various_studies}, we use a subset of the benchmarks for the same reasons.\footnote{Except for the function set study, were full coverage is key for a clear picture.} Wherever we have access to the original predictions from previous work, we re-evaluate them with our metric and, for a fair comparison, only use those samples present in both our and their evaluation.

\subsection{Results}
\label{subsec:results}

\cref{tab:main} shows our zero-shot and few-shot results for GRASP and GRASP-F on the main Wikidata and Freebase benchmarks using GPT-4.1 models \cite{gpt41}. In the zero-shot setting we reach very good performance overall.
The GPT-4.1 variant sets a new state-of-the-art on all Wikidata benchmarks, while coming close to or surpassing the best few-shot approaches on Freebase. The inclusion of feedback improves performance across the board: it thus helps if the model gets the chance to reassess and refine its own outputs once in a while. In the few-shot setting we only evaluate GPT-4.1 mini due to cost.
As expected, adding examples improves performance even further, with few-shot GRASP using GPT-4.1 mini surpassing zero-shot GRASP-F using GPT-4.1 on 3 out of 5 benchmarks. Interestingly, feedback does not seem to help as much in the few-shot setting, probably because the in-context examples already provide sufficient information in this case.

\subsection{Various Studies}
\label{subsec:various_studies}

\subsubsection{Model Provider Study} \cref{tab:component_provider} shows the results for GRASP using various base models.
We find that open-source models like Qwen2.5 72B and Qwen3 32B \cite{qwen25,qwen3} can reach performance equal to or better than commercial models like Gemini 2.0 Flash \cite{gemini2}. Most commercial models still outperform the open-source models, with GPT-4.1 and o4-mini \cite{o4mini} at the top. The reasoning models do not significantly outperform non-reasoning models, probably because we explicitly instruct GRASP to ``think'' before and after each step, which leads to reasoning-like behavior.

\begin{table}
\centering
\caption{Comparison of our approach with previous work. * marks previous work where we re-evaluated the original predictions on the same sample set and with the same exact metric as our models. All other results are taken from the respective papers.
\textsuperscript{\textdagger} marks results taken from SPINACH \cite{spinach}. Results in [brackets] report Hits@1 instead of F\textsubscript{1}-score. Few-shot results on SPINACH are missing because this benchmark does not provide a dedicated train set. Our best and second best models per benchmark are bold and underlined respectively.}
\begin{tabular}{l@{\hskip 0.25cm}cc@{\hskip 0.25cm}cccc}
\toprule
\multirow{2}{*}{\textbf{Approach}} & \multicolumn{2}{c}{\textbf{Freebase}} & \multicolumn{4}{c}{\textbf{Wikidata}} \\
 & CWQ & WQSP & QALD-10 & QALD-7 & SPINACH & WWQ \\
\midrule
\multicolumn{7}{l}{Fine-tuned}\\
\midrule
DeCAF \cite{decaf} & [70.4] & 78.8 & - & - & - \\
Pangu \cite{pangu} & - & 79.6 & - & - & - & - \\
FlexKBQA \cite{flexkbqa} & - & 60.6 & - & - & - & - \\
ChatKBQA \cite{chatkbqa} & 77.8 & 79.8 & - & - & - & - \\
RGR-KBQA \cite{rgrkbqa} & 76.6 & 80.7 & - & - & - & - \\
RoG \cite{rog} & 56.2 & 70.8 & - & - & - & - \\
RwT \cite{rwt} & 66.7 & 79.7 & - & - & - & - \\
SymAgent \cite{symagent} & 48.3 & 57.1 & - & - & - & - \\
LightPROF \cite{lightprof} & [59.3] & [83.8] & - & - & - & - \\
WikiSP \cite{wikisp} & - & - & - & - & 7.1\textsuperscript{\textdagger} & 71.9 \\
\midrule
\multicolumn{7}{l}{Few-shot}\\
\midrule
KB-BINDER \cite{kbbinder} & - & 74.4 & - & - & - & - \\
Interactive-KBQA \cite{interactivekbqa} & 49.1 & 71.2 & - & - & - & - \\
Pangu \cite{pangu} & - & 68.3 & - & - & - & - \\
KELDaR-rel \cite{keldar} & [44.2] & [76.7] & - & - & - & - \\
DoG \cite{dog} & [58.2] & [91.0] & - & - & - & - \\
\midrule
\multicolumn{7}{l}{Zero-shot}\\
\midrule
StructGPT \cite{structgpt} & - & [72.6] & - & - & - & - \\
ToG \cite{tog} & [69.5] & [82.6] & [54.7] & - & 7.2\textsuperscript{\textdagger} & - \\
SPINACH \cite{spinach} & - & - & 69.5 & 74.6 & 45.3 & 70.3 \\
*SPINACH \cite{spinach} & - & - & 71.0 & 72.9 & 39.2 & 68.3 \\
\midrule\midrule
\multicolumn{7}{l}{Zero-shot}\\
\midrule
GRASP \scriptsize GPT-4.1 mini & 23.5 & 46.1 & 66.7 & 66.2 & 32.1 & 71.5 \\
GRASP-F \scriptsize GPT-4.1 mini & 32.2 & 52.1 & 69.5 & 68.6 & 35.4 & 71.3 \\
GRASP \scriptsize GPT-4.1 & 44.2 & 52.1 & \textbf{72.5} & \underline{79.4} & \underline{40.8} & 75.3 \\
GRASP-F \scriptsize GPT-4.1 & 58.5 & 62.2 & \underline{70.6} & \textbf{81.5} & \textbf{42.6} & 75.2 \\
\midrule
\multicolumn{7}{l}{Few-shot using GPT-4.1 mini}\\
\midrule
\makecell[l]{GRASP \scriptsize similar 3-shot} & \underline{63.7} & \textbf{71.0} & 68.1 & 74.5 & - & \textbf{79.7} \\
\makecell[l]{GRASP-F \scriptsize similar 3-shot} & \textbf{65.6} & \underline{67.9} & 67.4 & 73.0 & - & \underline{78.3}
\end{tabular}

\label{tab:main}
\end{table}

\begin{table}[h]
\centering
\caption{Comparison of GRASP when using different base models. Reasoning models are evaluated on 100 samples to save costs and compute. The best and second best models per benchmark are bold and underlined respectively.}
\begin{tabular}{l@{\hskip 0.25cm}c@{\hskip 0.25cm}cc}
\toprule
\multirow{2}{*}{\textbf{Model}} & \textbf{Freebase} & \multicolumn{2}{c}{\textbf{Wikidata}} \\
 & WQSP & QALD-10 & WWQ \\
\midrule
\multicolumn{4}{l}{Using non-reasoning models}\\
\midrule
Gemini 2.0 Flash & 41.4 & 58.5 & 62.6 \\
Qwen2.5 72B & 40.8 & 61.7 & 68.7 \\
GPT-4.1 mini & 46.1 & 66.7 & \underline{71.5} \\
GPT-4.1 & \underline{52.1} & \underline{72.5} & \textbf{75.3} \\
\midrule
\multicolumn{4}{l}{Using reasoning models}\\
\midrule
Qwen3 32B & 39.2 & 50.3 & 61.3 \\
Gemini 2.5 Flash & 47.4 & 62.7 & 68.6 \\
o4-mini & \textbf{59.2} & \textbf{82.3} & 65.2
\end{tabular}

\label{tab:component_provider}
\end{table}

\vspace{-2mm}
\subsubsection{Function set study} We run Qwen2.5 72B on all benchmarks with each of our function sets. As expected, we find that the base function set \FNSET{$B$} without explicit search functions performs worst across all benchmarks. This is because the builtin search functionality in SPARQL on literals is limited compared to our search indices, often leads to errors due to high memory usage or timeouts, and is harder to use for the model. We find that overall the \FNSET{$S_E$} function set performs the best. We hypothesize that this is because it is easier to use than \FNSET{$S_A$} or \FNSET{$S_C$} while still allowing constrained search unlike \FNSET{$S$}.
See \cref{tab:component_fnsets} for full results.

\begin{table}[t]
\centering
\caption{Comparison of GRASP when using different function sets. The best and second best models per benchmark are bold and underlined respectively.}
\begin{tabular}{l@{\hskip 0.25cm}cc@{\hskip 0.25cm}cccc}
\toprule
\multirow{2}{*}{\textbf{Set}} & \multicolumn{2}{c}{\textbf{Freebase}} & \multicolumn{4}{c}{\textbf{Wikidata}} \\
 & CWQ & WQSP & Q-10 & Q-7 & SPIN. & WWQ \\
\midrule
\multicolumn{7}{l}{Using Qwen2.5 72B}\\
\midrule
\FNSET{$B$} & 6.5 & 19.8 & 43.8 & 52.6 & 17.3 & 55.7 \\
\FNSET{$S$} & \underline{23.9} & 36.8 & \underline{63.3} & \underline{73.0} & 27.4 & \textbf{68.7} \\
\FNSET{$S_C$} & 23.1 & 38.1 & \textbf{67.0} & 68.7 & 24.1 & 67.5 \\
\FNSET{$S_A$} & 16.2 & \underline{40.3} & 60.1 & 70.8 & \textbf{27.9} & \underline{68.4} \\
\FNSET{$S_E$} & \textbf{29.9} & \textbf{40.8} & 61.7 & \textbf{73.5} & \underline{27.6} & \textbf{68.7}
\end{tabular}

\label{tab:component_fnsets}
\vspace{-4mm}
\end{table}

\vspace{-2mm}

\subsubsection{Model Size Study} The Qwen2.5 and Qwen3 model families provide a range of different model sizes which we use to determine how model size affects performance with GRASP. As expected, we find that performance increases with model size. There also seems to be quite a jump in performance between the 7\B and 14\B models for Qwen2.5 and between the 4\B and 8\B models for Qwen3. The best model overall is Qwen2.5 72\B, though a model of comparable size is missing for Qwen3. See \cref{tab:component_modelsize} for full results.

\vspace{-2mm}
\subsubsection{Few-shot and Feedback Study} In addition to the results from \cref{tab:main}, we also run GPT-4.1 mini with random few-shot examples. As mentioned in \cref{subsec:results}, our results show that feedback generally improves performance if no examples are given but does not help as much, and sometimes even hurts, if there are examples. Also as expected, in most cases both types of examples improve performance, with similar examples helping more than random examples, especially on benchmarks with more similar train and test splits. We validate these findings with a separate set of evaluations using Qwen2.5 72B and observe similar patterns there. As a rule of thumb, one should generally use feedback for best performance, but it can be disabled if there are examples available, especially if they are known to be similar to expected future questions. See \cref{tab:component_extras} for full results.

\begin{table}[h]
\centering
\begin{minipage}[t]{0.43\textwidth}
\centering
\caption{Comparison of GRASP for different model sizes with Qwen2.5 (non-reasoning) and Qwen3 (reasoning). The best and second best models per benchmark are bold and underlined respectively.}
\begin{tabular}{l@{\hskip 0.25cm}c@{\hskip 0.25cm}cc}
\toprule
\multirow{2}{*}{\textbf{Size}} & \textbf{Freebase} & \multicolumn{2}{c}{\textbf{Wikidata}} \\
 & WQSP & QALD-10 & WWQ \\
\midrule
\multicolumn{4}{l}{Using Qwen2.5 models}\\
\midrule
7B & 24.1 & 39.3 & 37.0 \\
14B & 38.2 & 51.6 & 60.2 \\
32B & 29.9 & \underline{59.8} & 58.7 \\
72B & \textbf{40.8} & \textbf{61.7} & \textbf{68.7} \\
\midrule
\multicolumn{4}{l}{Using Qwen3 models}\\
\midrule
4B & 25.2 & 25.4 & 35.3 \\
8B & 33.6 & 46.2 & 51.5 \\
14B & 24.4 & 45.7 & 50.7 \\
32B & \underline{39.2} & 50.3 & \underline{61.3}
\end{tabular}

\label{tab:component_modelsize}
\end{minipage}\hfill%
\begin{minipage}[t]{0.55\textwidth}
\centering
\caption{Effect of feedback and few-shot examples on performance. Models using three similar examples are marked as \emph{sim}, while models using three random examples are marked as \emph{rand}. The best and second best models per benchmark are bold and underlined respectively.}
\begin{tabular}{l@{\hskip 0.25cm}c@{\hskip 0.25cm}cc}
\toprule
\multirow{2}{*}{\textbf{Variant}} & \textbf{Freebase} & \multicolumn{2}{c}{\textbf{Wikidata}} \\
 & WQSP & QALD-10 & WWQ \\
\midrule
\multicolumn{4}{l}{Using GPT-4.1 mini}\\
\midrule
GRASP & 46.1 & 66.7 & 71.5 \\
GRASP-F & 52.1 & \underline{69.1} & 71.3 \\
GRASP \scriptsize \emph{rand} & 57.4 & \textbf{70.8} & 70.5 \\
GRASP-F \scriptsize \emph{rand}& 59.0 & 67.3 & 71.2 \\
GRASP \scriptsize \emph{sim} & \textbf{71.0} & 67.8 & \textbf{79.7} \\
GRASP-F \scriptsize \emph{sim} & \underline{67.9} & 67.1 & \underline{78.3} \\
\midrule
\multicolumn{4}{l}{Using Qwen2.5 72B}\\
\midrule
GRASP & 40.8 & 61.7 & 68.7 \\
GRASP-F & 47.6 & 64.1 & 64.6 \\
GRASP \scriptsize \emph{rand} & 45.0 & 65.1 & 66.4 \\
GRASP-F \scriptsize \emph{rand} & 57.0 & 62.5 & 63.2 \\
GRASP \scriptsize \emph{sim} & 65.9 & 65.0 & 76.2 \\
GRASP-F \scriptsize \emph{sim} & 66.5 & 68.0 & 70.4
\end{tabular}

\label{tab:component_extras}
\end{minipage}
\end{table}

\subsubsection{Knowledge Graph Study} To show that our approach also generalizes to other, less commonly evaluated knowledge graphs, we collect or create benchmarks for DBLP, DBpedia, ORKG, OSM, and UniProt, and add two more benchmarks for Wikidata and one for Freebase. We evaluate GRASP-F using GPT-4.1 on all of them and show the results in \cref{tab:component_kgs}. Our approach performs well overall, except on OSM and UniProt (for reasons partly explained next).

\begin{table}[h!]
\centering
\caption{Performance of GRASP-F on other knowledge graphs and benchmarks. We evaluate on 50 randomly selected samples per benchmark.}
\begin{tabularx}{\textwidth}{llcX}
\toprule
\textbf{KG} & \textbf{Benchmark} & \textbf{F\textsubscript{1}-score} & \textbf{Description} \\
\midrule
\multicolumn{4}{l}{Using GPT-4.1}\\
\midrule
DBLP & DBLP-QuAD & 51.0 & KGQA benchmark over DBLP \cite{dblpquad} \\
DBLP & Examples & 50.3 & Examples for the RDF version of DBLP \cite{dblptgdk,dblptgdkexamples} \\
DBpedia & SimpleQ & 53.5 & Migration of Freebase SimpleQ to DBpedia \cite{simplequestionsdbpedia} \\
DBpedia & LC-QuAD & 48.6 & Template-based SPARQL queries over DBpedia \cite{lcquad} \\
DBpedia & QALD-7 & 60.3 & English DBpedia subset of QALD-7 \cite{qald7} \\
DBpedia & QALD-9 & 65.2 & English subset of QALD-9 \cite{qald9}\\
Freebase & SimpleQ & 59.8 & Questions answerable by a single triple \cite{simplequestionsfreebase} \\
ORKG & SciQA & 52.2 & KGQA benchmark over ORKG \cite{sciqa} \\ 
OSM & Examples & 17.3 & Examples from the OSM website \cite{osmexamples} \\
UniProt & Examples & 20.7 & Examples from the UniProt website \cite{uniprotexamples} \\
Wikidata & SimpleQ & 67.6 & Migration of Freebase SimpleQ to Wikidata \cite{simplequestions} \\
Wikidata & LC-QuAD 2.0 & 62.1 & Wikidata variant of the LC-QuAD successor \cite{lcquad2}
\end{tabularx}

\label{tab:component_kgs}
\end{table}

\subsection{Error Analysis}
\label{subsec:error_analysis}

We manually inspect a random subset of the results of our best variant (GRASP-F using GPT-4.1) for each benchmark.
For almost all queries, the model performs sensible function calls, which are suited to lead to the correct query. There is no dominant source of error across graphs, but each graph has some quirks that are hard for the model to handle. Here are three examples:\\[1mm]
1. Wikidata potentially contains the same information in different ways. Sometimes, a complex information is captured by a single IRI, such as \href{https://www.wikidata.org/entity/Q93306595}{\emph{Mayor of Germasogeia Municipality Elections (wd:Q93306595)}}. Sometimes, such information uses a highly non-trivial combination of properties like \emph{p:2139, psv:P2139, pq:P585}. Our model sometimes commits to the wrong alternatives, which then gives suboptimal or no results. This is a hard problem, also for expert users.\\[1mm]
2. Several of the DBLP queries involve citations. However, there is no triple that directly connects the DBLP ID for a paper with a cited or citing paper or with an object that represents a citation. Instead, there is a predicate \emph{dblp:omid} that connects the DBLP ID to another kind of ID (called OMID), which in turn is connected to an object representing a citation. Our models desperately look for this connection but don't find it. This, too, is hard for experts.\\[1mm]
3. Several of the OSM examples involve finding members of OSM ways or relations. The obvious predicate \emph{ogc:sfContains} does not work, instead the predicates \emph{osmrel:member} and \emph{osmway:member} are needed. Our models do encounter these predicates during their search, but do not follow through with them. In general, the OSM and UniProt queries are considerable more complex.\\[1mm]
Independently of the graph, our model tends to forgo the property search and hallucinate properties, despite unsuccessful query executions. 
We tried to mitigate this behavior by returning error messages to the model if it uses IRIs within \FN{EXE} that were not part of a prior function call result, and got some initial positive results that this indeed helps; exploring this further is an interesting direction for future work.
Another kind of error are ambiguous questions, where ground truth and generated SPARQL query are both reasonable but different. Similarly, ground truth and generated query sometimes only differ in the LIMIT clause. In both cases, the model gets an unfairly low F\textsubscript{1}-score. It would be desirable to develop metrics that are more robust against such errors.

\section{Conclusion}
\label{sec:conclusion}

We present GRASP, a new approach for zero-shot generation of SPARQL queries over arbitrary knowledge graphs. In an extensive evaluation across many graphs, GRASP performs well across the board. On Wikidata, it achieves state-of-the-art in the zero-shot setting, and comes close to few-shot or fine-tuned models on Freebase. We investigated GRASP's individual components (function set, language model, feedback mechanism, example incorporation) in detail. It turns out that search functions that are both context-sensitive and easy to use work best. GRASP works well with both commercial and open-source models, with a slight edge for GPT-4.1. Both examples and feedback help to improve performance.


\paragraph*{Supplemental Material Statement:} All of our code and data (e.g. knowledge graph indices, benchmarks, model outputs, etc.) is openly available via \REPRO. This includes a web app for evaluating and comparing models.

\subsubsection{Acknowledgments}
Funded by the Deutsche Forschungsgemeinschaft (DFG, German Research Foundation) – Project-ID 499552394 – SFB 1597.

\bibliographystyle{splncs04}
\bibliography{sparql-qa}

\begin{thebibliography}{10}
\providecommand{\url}[1]{\texttt{#1}}
\providecommand{\urlprefix}{URL }
\providecommand{\doi}[1]{https://doi.org/#1}

\bibitem{dblptgdk}
Ackermann, M.R., Bast, H., Beckermann, B.M., Kalmbach, J., Neises, P.,
  Ollinger, S.: The dblp knowledge graph and {SPARQL} endpoint. {TGDK}
  \textbf{2}(2),  3:1--3:23 (2024). \doi{10.4230/TGDK.2.2.3},
  \url{https://doi.org/10.4230/TGDK.2.2.3}

\bibitem{lightprof}
Ao, T., Yu, Y., Wang, Y., Deng, Y., Guo, Z., Pang, L., Wang, P., Chua, T.,
  Zhang, X., Cai, Z.: {LightPROF}: A lightweight reasoning framework for large
  language model on knowledge graph. In: {AAAI}. pp. 23424--23432. {AAAI} Press
  (2025)

\bibitem{sciqa}
Auer, S., Barone, D.A., Bartz, C., Cortes, E.G., Jaradeh, M.Y., Karras, O.,
  Koubarakis, M., Mouromtsev, D., Pliukhin, D., Radyush, D., Shilin, I.,
  Stocker, M., Tsalapati, E.: The {SciQA} scientific question answering
  benchmark for scholarly knowledge. {Scientific Reports}  \textbf{13}(1),
  ~7240 (2023)

\bibitem{dbpedia}
Auer, S., Bizer, C., Kobilarov, G., Lehmann, J., Cyganiak, R., Ives, Z.G.:
  {DBpedia}: A nucleus for a web of open data. In: {ISWC/ASWC}. Lecture Notes
  in Computer Science, vol.~4825, pp. 722--735. Springer (2007)

\bibitem{orkg2023}
Auer, S., Barone, D.A.C., Bartz, C., Cortes, E.G., Jaradeh, M.Y., Karras, O.,
  Koubarakis, M., Mouromtsev, D., Pliukhin, D., Radyush, D., Shilin, I.,
  Stocker, M., Tsalapati, E.: {SciQA} benchmark: Dataset and {RDF} dump (2023).
  \doi{10.5281/zenodo.7744048}, \url{https://doi.org/10.5281/zenodo.7744048}

\bibitem{simplequestionsdbpedia}
Azmy, M., Shi, P., Lin, J., Ilyas, I.F.: Farewell {Freebase}: Migrating the
  {SimpleQuestions} dataset to {DBpedia}. In: {COLING}. pp. 2093--2103.
  Association for Computational Linguistics (2018)

\bibitem{dblpquad}
Banerjee, D., Awale, S., Usbeck, R., Biemann, C.: {DBLP-QuAD}: A question
  answering dataset over the {DBLP} scholarly knowledge graph. In: {BIR@ECIR}.
  {CEUR} Workshop Proceedings, vol.~3617, pp. 37--51. CEUR-WS.org (2023)

\bibitem{osm2rdf}
Bast, H., Brosi, P., Kalmbach, J., Lehmann, A.: An efficient {RDF} converter
  and {SPARQL} endpoint for the complete {OpenStreetMap} data. In:
  {SIGSPATIAL/GIS}. pp. 536--539. {ACM} (2021)

\bibitem{qlever}
Bast, H., Buchhold, B.: {QLever}: A query engine for efficient {SPARQL}+{Text}
  search. In: {CIKM}. pp. 647--656. {ACM} (2017)

\bibitem{qleverautocompletion}
Bast, H., Kalmbach, J., Klumpp, T., Kramer, F., Schnelle, N.: Efficient and
  effective {SPARQL} autocompletion on very large knowledge graphs. In: {CIKM}.
  pp. 2893--2902. {ACM} (2022)

\bibitem{webquestions}
Berant, J., Chou, A., Frostig, R., Liang, P.: Semantic parsing on {Freebase}
  from question-answer pairs. In: {EMNLP}. pp. 1533--1544. {ACL} (2013)

\bibitem{freebase}
Bollacker, K.D., Evans, C., Paritosh, P.K., Sturge, T., Taylor, J.: Freebase: a
  collaboratively created graph database for structuring human knowledge. In:
  {SIGMOD} Conference. pp. 1247--1250. {ACM} (2008)

\bibitem{dblptgdkexamples}
{DBLP GitHub}: Wiki page for {TGDK 2024} paper.
  \url{https://github.com/dblp/kg/wiki/Paper-TGDK-2024}, accessed: 2025-05-07

\bibitem{simplequestions}
Diefenbach, D., Tanon, T.P., Singh, K.D., Maret, P.: Question answering
  benchmarks for {Wikidata}. In: {ISWC} (Posters, Demos {\&} Industry Tracks).
  {CEUR} Workshop Proceedings, vol.~1963. CEUR-WS.org (2017)

\bibitem{faiss}
Douze, M., Guzhva, A., Deng, C., Johnson, J., Szilvasy, G., Mazaré, P.E.,
  Lomeli, M., Hosseini, L., Jégou, H.: The {Faiss} library  (2024)

\bibitem{lcquad2}
Dubey, M., Banerjee, D., Abdelkawi, A., Lehmann, J.: {LC-QuAD 2.0}: A large
  dataset for complex question answering over {Wikidata} and {DBpedia}. In:
  {ISWC} {(2)}. Lecture Notes in Computer Science, vol. 11779, pp. 69--78.
  Springer (2019)

\bibitem{rgrkbqa}
Feng, T., He, L.: {RGR-KBQA:} generating logical forms for question answering
  using knowledge-graph-enhanced large language model. In: {COLING}. pp.
  3057--3070. Association for Computational Linguistics (2025)

\bibitem{uniprot-rdf}
Garcia, L., Bolleman, J., Gehant, S., Redaschi, N., Martin, M., Bateman, A.,
  Magrane, M., Orchard, S., Raj, S., Ahmad, S., Alpi, E., Bowler, E., Britto,
  R., Bursteinas, B., Bye-A-Jee, H., Dogan, T., Garmiri, P., Georghiou, G.,
  Gonzales, L., Hatton-Ellis, E., Ignatchenko, A., Insana, G., Ishtiaq, R.,
  Joshi, V., Jyothi, D., Luo, J., Lussi, Y., MacDougall, A., Mahmoudy, M.,
  Nightingale, A., Oliveira, C., Onwubiko, J., Poddar, V., Pundir, S., Qi, G.,
  Rifaioglu, A., Rice, D., Saidi, R., Speretta, E., Turner, E., Tyagi, N.,
  Vasudev, P., Volynkin, V., Warner, K., Watkins, X., Zaru, R., Zellner, H.,
  Bridge, A., Breuza, L., Coudert, E., Lieberherr, D., Pedruzzi, I., Poux, S.,
  Pruess, M., Aimo, L., Argoud-Puy, G., Auchincloss, A., Axelsen, K., Bansal,
  P., Baratin, D., Batista~Neto, T., Blatter, M.C., Boutet, E., Casals-Casas,
  C., Cuche, B., De~Castro, E., Estreicher, A., Famiglietti, L., Feuermann, M.,
  Gasteiger, E., Gerritsen, V., Gos, A., Gruaz, N., Hinz, U., Hulo, C.,
  Hyka-Nouspikel, N., Jungo, F., Kerhornou, A., Lemercier, P., Lombardot, T.,
  Masson, P., Morgat, A., Pilbout, S., Pozzato, M., Rivoire, C., Sigrist, C.,
  Sundaram, S., Wu, C., Arighi, C., Huang, H., McGarvey, P., Natale, D.,
  Arminski, L., Chen, C., Chen, Y., Garavelli, J., Laiho, K., Ross, K.,
  Vinayaka, C.R., Wang, Q., Wang, Y., Yeh, L.S., Zhang, J., Consortium, U.:
  {FAIR} adoption, assessment and challenges at {UniProt}. Scientific Data
  \textbf{6}(1), ~175 (Sep 2019). \doi{10.1038/s41597-019-0180-9},
  \url{https://doi.org/10.1038/s41597-019-0180-9}

\bibitem{gemini2}
{Google DeepMind}: Introducing {Gemini 2.0}: our new {AI} model for the agentic
  era (2024),
  \url{https://blog.google/technology/google-deepmind/google-gemini-ai-update-december-2024/},
  accessed: 2025-05-11

\bibitem{pangu}
Gu, Y., Deng, X., Su, Y.: Don't generate, discriminate: A proposal for
  grounding language models to real-world environments. In: {ACL} {(1)}. pp.
  4928--4949. Association for Computational Linguistics (2023)

\bibitem{simplequestionsfreebase}
He, X., Golub, D.: Character-level question answering with attention. In:
  {EMNLP}. pp. 1598--1607. The Association for Computational Linguistics (2016)

\bibitem{orkg}
Jaradeh, M.Y., Oelen, A., Farfar, K.E., Prinz, M., D'Souza, J., Kismih{\'{o}}k,
  G., Stocker, M., Auer, S.: Open research knowledge graph: Next generation
  infrastructure for semantic scholarly knowledge. In: {K-CAP}. pp. 243--246.
  {ACM} (2019)

\bibitem{structgpt}
Jiang, J., Zhou, K., Dong, Z., Ye, K., Zhao, X., Wen, J.: {StructGPT}: A
  general framework for large language model to reason over structured data.
  In: {EMNLP}. pp. 9237--9251. Association for Computational Linguistics (2023)

\bibitem{mxbai}
Lee, S., Shakir, A., Koenig, D., Lipp, J.: Open source strikes bread - new
  fluffy embedding model (2024),
  \url{https://www.mixedbread.ai/blog/mxbai-embed-large-v1}

\bibitem{keldar}
Li, Y., Song, D., Zhou, C., Tian, Y., Wang, H., Yang, Z., Zhang, S.: A
  framework of knowledge graph-enhanced large language model based on question
  decomposition and atomic retrieval. In: {EMNLP} (Findings). pp. 11472--11485.
  Association for Computational Linguistics (2024)

\bibitem{flexkbqa}
Li, Z., Fan, S., Gu, Y., Li, X., Duan, Z., Dong, B., Liu, N., Wang, J.:
  {FlexKBQA}: A flexible {LLM}-powered framework for few-shot knowledge base
  question answering. In: {AAAI}. pp. 18608--18616. {AAAI} Press (2024)

\bibitem{symagent}
Liu, B., Zhang, J., Lin, F., Yang, C., Peng, M., Yin, W.: {SymAgent}: A
  neural-symbolic self-learning agent framework for complex reasoning over
  knowledge graphs. In: {WWW}. pp. 98--108. {ACM} (2025)

\bibitem{spinach}
Liu, S., Semnani, S.J., Triedman, H., Xu, J., Zhao, I.D., Lam, M.S.: {SPINACH:}
  {SPARQL}-based information navigation for challenging real-world questions.
  In: {EMNLP} (Findings). pp. 15977--16001. Association for Computational
  Linguistics (2024)

\bibitem{chatkbqa}
Luo, H., E, H., Tang, Z., Peng, S., Guo, Y., Zhang, W., Ma, C., Dong, G., Song,
  M., Lin, W., Zhu, Y., Luu, A.T.: {ChatKBQA}: A generate-then-retrieve
  framework for knowledge base question answering with fine-tuned large
  language models. In: {ACL} (Findings). pp. 2039--2056. Association for
  Computational Linguistics (2024)

\bibitem{rog}
Luo, L., Li, Y., Haffari, G., Pan, S.: Reasoning on graphs: Faithful and
  interpretable large language model reasoning. In: {ICLR}. OpenReview.net
  (2024)

\bibitem{dog}
Ma, J., Gao, Z., Chai, Q., Sun, W., Wang, P., Pei, H., Tao, J., Song, L., Liu,
  J., Zhang, C., Cui, L.: Debate on graph: A flexible and reliable reasoning
  framework for large language models. In: {AAAI}. pp. 24768--24776. {AAAI}
  Press (2025)

\bibitem{gpt41}
{OpenAI}: Introducing {GPT-4.1} in the {API} (April 2025),
  \url{https://openai.com/index/gpt-4-1/}, accessed: 2025-05-11

\bibitem{o4mini}
{OpenAI}: {OpenAI o3} and o4-mini system card. Tech. rep., OpenAI (April 2025),
  \url{https://cdn.openai.com/pdf/2221c875-02dc-4789-800b-e7758f3722c1/o3-and-o4-mini-system-card.pdf},
  accessed: 2025-05-11

\bibitem{osmexamples}
{OpenStreetMap Website}: {QLever} example queries.
  \url{https://wiki.openstreetmap.org/wiki/QLever/Example_queries}, accessed:
  2025-05-07

\bibitem{kbbinder}
Patidar, M., Sawhney, R., Singh, A.K., Chatterjee, B., Mausam, Bhattacharya,
  I.: Few-shot transfer learning for knowledge base question answering: Fusing
  supervised models with in-context learning. In: {ACL} {(1)}. pp. 9147--9165.
  Association for Computational Linguistics (2024)

\bibitem{qwen3}
{Qwen Team}: {Qwen3} technical report (2025),
  \url{https://arxiv.org/abs/2505.09388}

\bibitem{sentencetransformers}
Reimers, N., Gurevych, I.: {Sentence-BERT}: Sentence embeddings using siamese
  {BERT}-networks. In: Proceedings of the 2019 Conference on Empirical Methods
  in Natural Language Processing. Association for Computational Linguistics (11
  2019), \url{https://arxiv.org/abs/1908.10084}

\bibitem{rwt}
Shen, T., Wang, J., Zhang, X., Cambria, E.: Reasoning with trees: Faithful
  question answering over knowledge graph. In: {COLING}. pp. 3138--3157.
  Association for Computational Linguistics (2025)

\bibitem{tog}
Sun, J., Xu, C., Tang, L., Wang, S., Lin, C., Gong, Y., Ni, L.M., Shum, H.,
  Guo, J.: {Think-on-Graph}: Deep and responsible reasoning of large language
  model on knowledge graph. In: {ICLR}. OpenReview.net (2024)

\bibitem{cwq}
Talmor, A., Berant, J.: The web as a knowledge-base for answering complex
  questions. In: {NAACL-HLT}. pp. 641--651. Association for Computational
  Linguistics (2018)

\bibitem{lcquad}
Trivedi, P., Maheshwari, G., Dubey, M., Lehmann, J.: {LC-QuAD}: A corpus for
  complex question answering over knowledge graphs. In: {ISWC} {(2)}. Lecture
  Notes in Computer Science, vol. 10588, pp. 210--218. Springer (2017)

\bibitem{uniprotexamples}
{UniProt Website}: {SPARQL} example queries.
  \url{https://sparql.uniprot.org/.well-known/sparql-examples}, accessed:
  2025-05-07

\bibitem{qald9}
Usbeck, R., Gusmita, R.H., Ngomo, A.N., Saleem, M.: 9th challenge on question
  answering over linked data {(QALD-9)} (invited paper). In:
  Semdeep/NLIWoD@ISWC. {CEUR} Workshop Proceedings, vol.~2241, pp. 58--64.
  CEUR-WS.org (2018)

\bibitem{qald7}
Usbeck, R., Ngomo, A.N., Haarmann, B., Krithara, A., R{\"{o}}der, M.,
  Napolitano, G.: 7th open challenge on question answering over linked data
  {(QALD-7)}. In: SemWebEval@ESWC. Communications in Computer and Information
  Science, vol.~769, pp. 59--69. Springer (2017)

\bibitem{qald10}
Usbeck, R., Yan, X., Perevalov, A., Jiang, L., Schulz, J., Kraft, A., Möller,
  C., Huang, J., Reineke, J., Ngomo, A.C.N., Saleem, M., Both, A.: {QALD-10}
  – the 10th challenge on question answering over linked data: Shifting from
  {DBpedia} to {Wikidata} as a {KG} for {KGQA}. Semantic Web  \textbf{15}(6),
  2193--2207 (2024). \doi{10.3233/SW-233471},
  \url{https://journals.sagepub.com/doi/abs/10.3233/SW-233471}

\bibitem{wikidata}
Vrandecic, D., Kr{\"{o}}tzsch, M.: Wikidata: a free collaborative
  knowledgebase. Commun. {ACM}  \textbf{57}(10),  78--85 (2014)

\bibitem{interactivekbqa}
Xiong, G., Bao, J., Zhao, W.: {Interactive-KBQA}: Multi-turn interactions for
  knowledge base question answering with large language models. In: {ACL}
  {(1)}. pp. 10561--10582. Association for Computational Linguistics (2024)

\bibitem{wikisp}
Xu, S., Liu, S., Culhane, T., Pertseva, E., Wu, M., Semnani, S.J., Lam, M.S.:
  Fine-tuned {LLM}s know more, hallucinate less with few-shot
  sequence-to-sequence semantic parsing over {Wikidata}. In: {EMNLP}. pp.
  5778--5791. Association for Computational Linguistics (2023)

\bibitem{qwen25}
Yang, A., Yang, B., Zhang, B., Hui, B., Zheng, B., Yu, B., Li, C., Liu, D.,
  Huang, F., Wei, H., et~al.: {Qwen2.5} technical report. arXiv preprint
  arXiv:2412.15115  (2024)

\bibitem{react}
Yao, S., Zhao, J., Yu, D., Du, N., Shafran, I., Narasimhan, K.R., Cao, Y.:
  {ReAct}: Synergizing reasoning and acting in language models. In: {ICLR}.
  OpenReview.net (2023)

\bibitem{webqsp}
Yih, W., Richardson, M., Meek, C., Chang, M., Suh, J.: The value of semantic
  parse labeling for knowledge base question answering. In: {ACL} {(2)}. The
  Association for Computer Linguistics (2016)

\bibitem{decaf}
Yu, D., Zhang, S., Ng, P., Zhu, H., Li, A.H., Wang, J., Hu, Y., Wang, W.Y.,
  Wang, Z., Xiang, B.: {DecAF}: Joint decoding of answers and logical forms for
  question answering over knowledge bases. In: {ICLR}. OpenReview.net (2023)

\end{thebibliography}


\end{document}